\documentclass[runningheads]{llncs}
\usepackage{graphicx}
\usepackage{amsmath,amssymb} % define this before the line numbering.
\usepackage{color}
\begin{document}
\title{Approach for Video Classification with Multi-label on YouTube-8M Dataset} 
% Replace with your title

\titlerunning{Approach for Video Classification with Multi-label on YouTube-8M Dataset}
% Replace with a meaningful short version of your title
%
%\author{First Author\inst{1}\orcidID{0000-0002-8511-8838} \and
%Second Author\inst{2,3}\orcidID{1111-2222-3333-4444} \and
%Third Author\inst{3}\orcidID{2222--3333-4444-5555}}

\newcommand*\samethanks[1][\value{footnote}]{\footnotemark[#1]}
\author{Kwangsoo Shin\thanks{These two authors contributed equally}\orcidID{0000-0002-8511-8838}, Junhyeong Jeon\samethanks\orcidID{0000-0002-8151-8375}, Seungbin Lee\orcidID{0000-0002-1917-0224}, Boyoung Lim, Minsoo Jeong, Jongho Nang}

%
%Please write out author names in full in the paper, i.e. full given and family names. 
%If any authors have names that can be parsed into FirstName LastName in multiple ways, please include the correct parsing, in a comment to the volume editors:
%\index{Lastnames, Firstnames}
%(Do not uncomment it, because you may introduce extra index items if you do that, we will use scripts for introducing index entries...)
\authorrunning{Kwangsoo Shin, Junhyeong Jeon, et al.}
% Replace with shorter version of the author list. If there are more authors than fits a line, please use A. Author et al.
%

\institute{
Department of Computer Science and Engineering, Sogang University \\
\email{\{ksshin,junhyeong.jeon,mercileesb,bylim,msjeong,jhnang\}@sogang.ac.kr}\\
\url{http://mmlab.sogang.ac.kr}}
\maketitle              % typeset the header of the contribution
\begin{abstract}
Video traffic is increasing at a considerable rate due to the spread of personal media and advancements in media technology.
Accordingly, there is a growing need for techniques to automatically classify moving images.
This paper use NetVLAD and NetFV models and the Huber loss function for video classification problem and YouTube-8M dataset to verify the experiment.
We tried various attempts according to the dataset and optimize hyperparameters, ultimately obtain a GAP score of 0.8668.

\end{abstract}

\section{Introduction}
\quad
Video traffic from video sites such as YouTube has increased in recent years. 
The growth of personal media through technological development is particularly remarkable.
With the development of smartphones, media is now brought to the consumer’s hand, and individuals are no longer only consumers of multimedia but are now producers as well.
This trend may be confirmed by internet traffic statistics and other global data.
As a result, it is becoming increasingly difficult for consumers to identify desirable media.
Accordingly, there is a growing need for techniques to recommend videos or automatically classify subjects.
Much effort has been made to process video.
Many recent advancements in artificial neural networks have been applied to video processing in an attempt to understand each frame of the video using a convolutional neural network (CNN) \cite{kim2014convolutional}.
Other methods used include VLAD \cite{jegou2010aggregating} for processing time series data, recurrent neural network network (RNN) series and long short-term memory (LSTM) \cite{hochreiter1997long} or GRU \cite{cho2014properties} for processing time series data.
The skipLSTM and skipGRU \cite{campos2018skip}, which add a skip connection to the RNN network, have been proven effective ways to process time series data. 
The present paper uses NetVLAD and NetFV, which are known as effective methods for video processing, to find the optimal network by adjusting various hyperparameters used in the network.
A single model was sought to solve the problem rather than an ensemble technique.
In this process, YouTube-8M dataset was used, and a GAP rating of 0.8668 was obtained for the test set.

\section{Methods and Materials}

\subsection{Dataset}
\quad
The total number of video in YouTube-8M dataset is 6.1 million.
The training set consists of 3.9 million videos.
The test and validation sets are each 1.1 million.
All video has an average of three labels, and each label is composed of 3,862 multi-labels.
Every video is between 120 and 500 seconds in length.
This paper use frame-level and audio features.
The frame-level features are 1,024-dimensional vectors in which selects one frame per second in video and extracts through Inception V3 model.
The audio features are extracted with 128-dimensional vectors drawn through a VGG-inspired acoustic model.

\subsection{Models}
\quad
This paper used NetVLAD \cite{arandjelovic2016netvlad} and NetFV \cite{perronnin2007fisher}, which were the most successful models used by Willow, the first place–winning team of the YouTube-8M video understanding challenge \cite{2017arXiv170606905M}.
A hyperparameter was identified to match the 2nd YouTube-8M video understanding challenge limit (model size < 1 GB).
NetVLAD and NetFV model uses integrated frame-level features and audio features.

\subsection{Loss Function}
\quad
This paper used the Huber loss function \cite{huber2011robust}, which was used by SNUVL X SKT when the team earned 8th place in the YouTube-8M video understanding challenge \cite{2017arXiv170607960N}.
The Huber loss combines L2 loss and L1 loss as shown in Equation~\ref{align:huber}.
As the YouTube-8M dataset is substantially imbalanced by a label, the Huber loss function was used to somewhat reduce the noise.

\begin{align}
  L_\delta (a) = \delta^2\left(\sqrt{1+(a/\delta)^2}-1\right)
\label{align:huber}
\end{align}

\subsection{Evaluation Metric}
\quad
In this paper, the global average precision (GAP) is used as an evaluation method.
The GAP is calculated with the top N predictions sorted by confidence score as shown in Equation~\ref{align:gap}.
\begin{align}
  GAP = \sum\limits_{i=1}^N p(i)\Delta r(i)
\label{align:gap}
\end{align}
\quad
In Equation~\ref{align:gap}, $p(i)$ is the precision and $r(i)$ is the recall.
In the 2nd YouTube-8M video understanding challenge, N is set to 20 and this paper is calculated accordingly also.

\section{Experiments}
\quad
The following experiments were performed: performance comparison of epoch, performance comparison by learning rate, performance comparison of modified dataset preprocessing.

\subsection{Epoch}
\label{sec:experimentepoch}
\begin{figure}
\centering
\includegraphics[height=6.5cm]{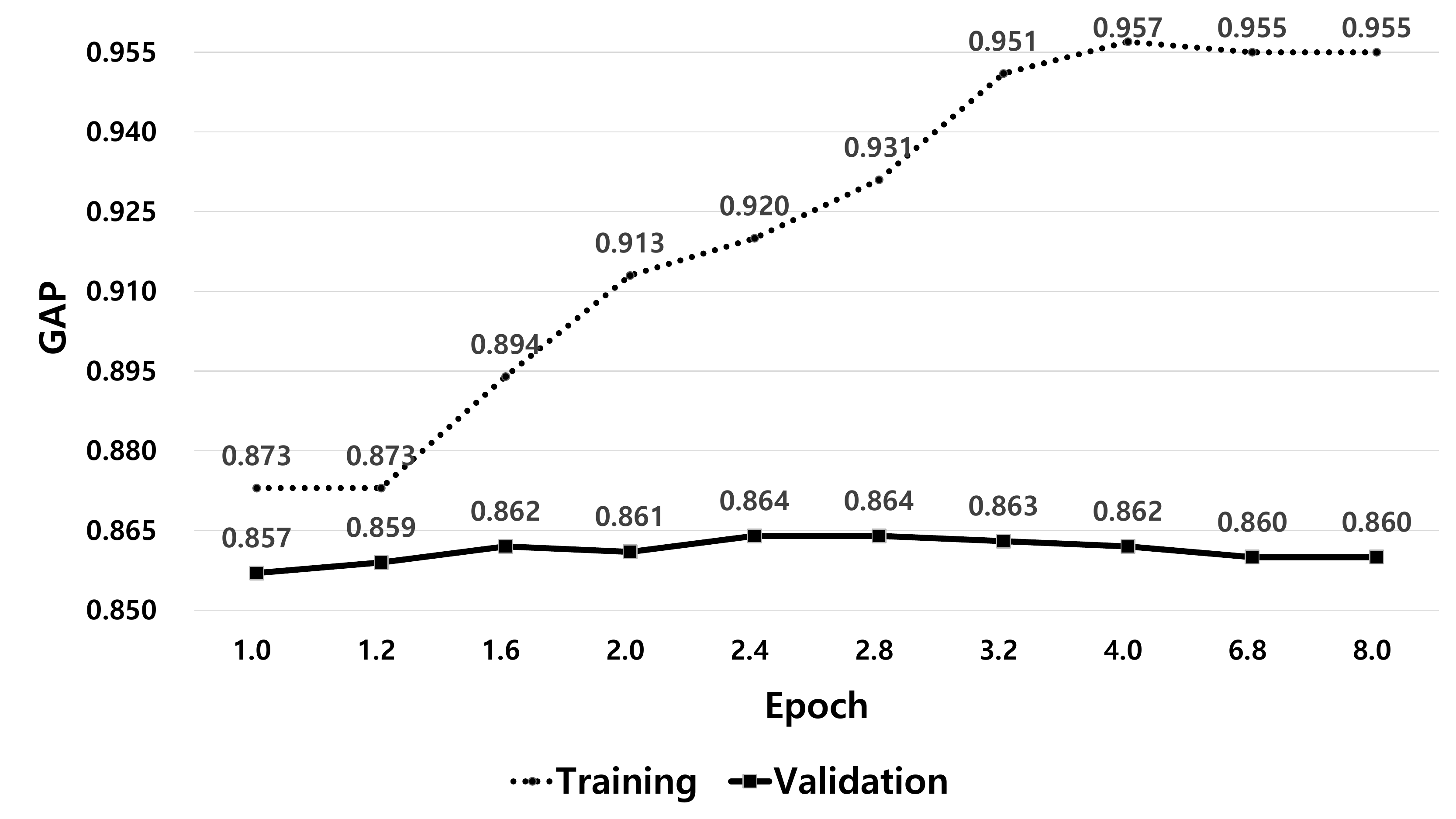}
\caption{\textbf{GAP per epoch curve.} The dotted line represents GAP per epoch in the training set; the solid line is GAP per epoch in the validation dataset. As the epoch increases, the training GAP curve also increases. However, the validation GAP curve shows a trend of declining after about 2.5 epochs.}
\label{fig:epoch}
\end{figure}

\quad
Of particular interest is the evaluation performed of each validation dataset for each epoch.
Usually, dozens of epochs are trained in the image training process.
However, that was not necessary for the YouTube-8M dataset.
The validation results for each epoch are shown in Fig.~\ref{fig:epoch}.
In the case of the YouTube-8M dataset, the training of one epoch was performed in approximately 25,000 steps when setting the batch size was 80 and used 2-GPU (total 160-batch).
Although the GAP difference was slight, it was possible to observe an optimal training performance between approximately 2.5 and 3 epochs.
In addition, the GAP for the training set increased in the additional training, but for the validation set decreased.
Similar trends were found for various parameters of various models. 
In this way, it was discovered that not many epochs were needed in the training process of this dataset, thus training was completed after 2.5 epochs.

\subsection{Learning Rate}
\begin{figure}
\centering
\includegraphics[height=6.5cm]{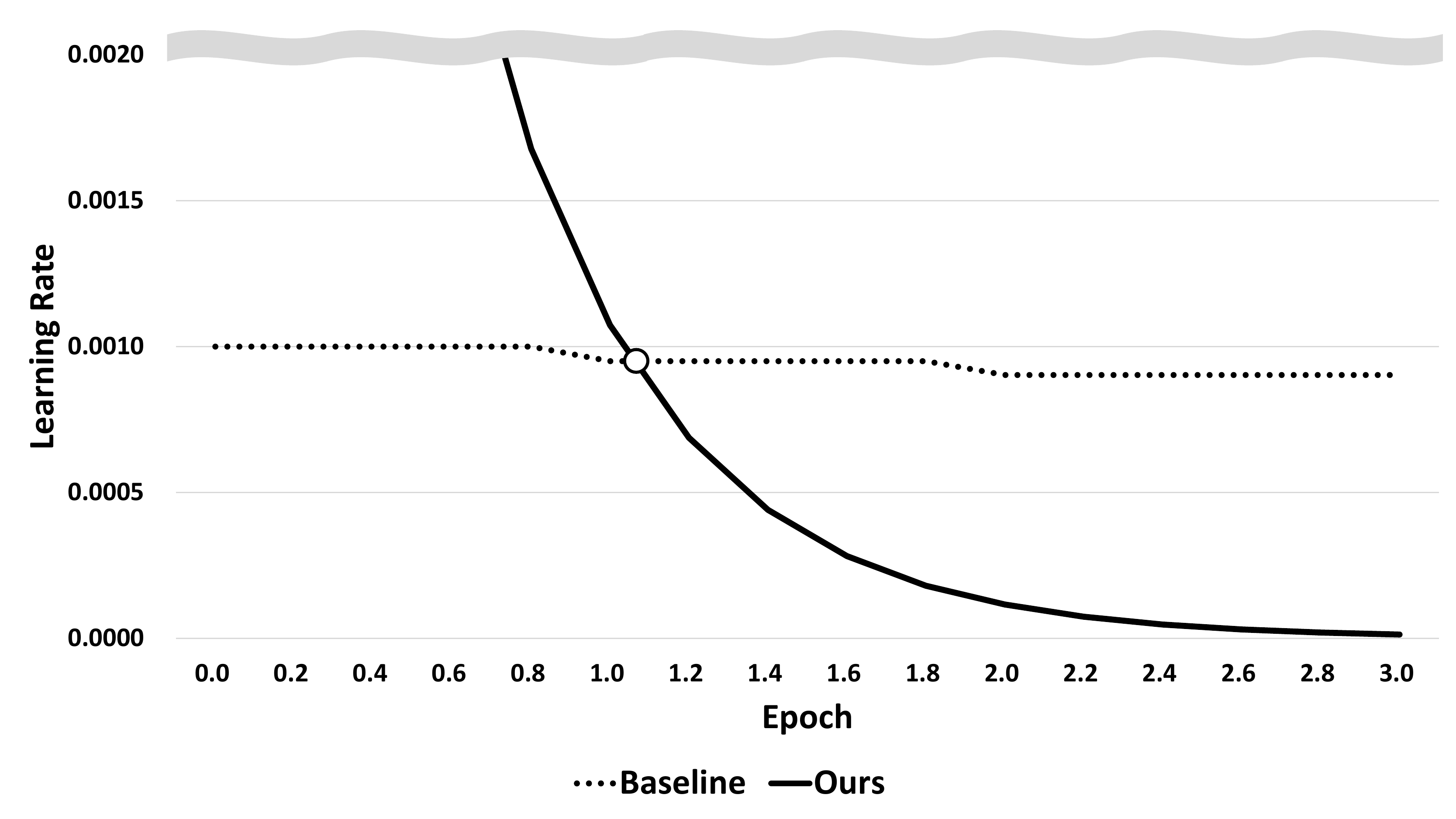}
\caption{\textbf{Comparison of baseline and experimental learning rates.} The initial learning rate of the present method was set at 10 times that of the baseline and the decay per epoch at 1/10 the baseline. The learning rate decay was modified so that the learning rate when the first epoch passed was similar to that of the baseline.}
\label{fig:learningratechange}
\end{figure}

\quad
The epoch experiment revealed that overfitting of the training set occurs when the model continues to train more than 3 epochs. 
This paper resolves this problem by adjusting the learning rate to be more effective.
In the early part of training, it is good to provide a relatively high learning rate to ensure quick training.
Conversely, at the end of the training, the learning rate should be decreased. 
Thus, the experiment began with a high learning rate, which was set to diminish over time.
The learning rate decay per epoch was set to 1/10 of the baseline.
It also increased the initial learning rate by 10 times that of the baseline.
These ways have helped reduce overfitting.
The learning rate decay was 0.8, for the purpose of keeping the learning rate of the two methods similar when the first epoch is complete.
So that the learning rate was greater than the baseline even if the training data is in the latter half.
This ensured training about these data.
The learning rate change at this method is shown in Fig.~\ref{fig:learningratechange}.
The resulting performance improvement is shown in Table~\ref{table:learningrate}.

\setlength{\tabcolsep}{4pt}
\begin{table}
\begin{center}
\caption{\textbf{Hyperparameters and its GAP score.} Hyperparameters were modified. GAP increased by 0.002.}
\label{table:learningrate}
\begin{tabular}{lll}
\hline\noalign{\smallskip}
Hyperparameter & Baseline & Ours \\
\noalign{\smallskip}
\hline
\noalign{\smallskip}
Initial learning rate & 0.001 & 0.01 \\
Learning rate decay & 0.95 & 0.80 \\
Learning rate decay per epoch & 1.0 & 0.1 \\
\noalign{\smallskip}
\hline
\noalign{\smallskip}
GAP & {\bf 0.864} & {\bf 0.866} \\
\hline
\end{tabular}
\end{center}
\end{table}
\setlength{\tabcolsep}{1.4pt}

\subsection{Data Preprocessing}
\quad
There was an attempt to improve performance through dataset modifications. First, the imbalance of the dataset was identified and addressed. Second, the false values of the results obtained were analyzed by validating the data trained with the default training set.

\subsubsection{Overfit to Non-dominant Pattern}
\begin{figure}
\centering
\includegraphics[height=6cm]{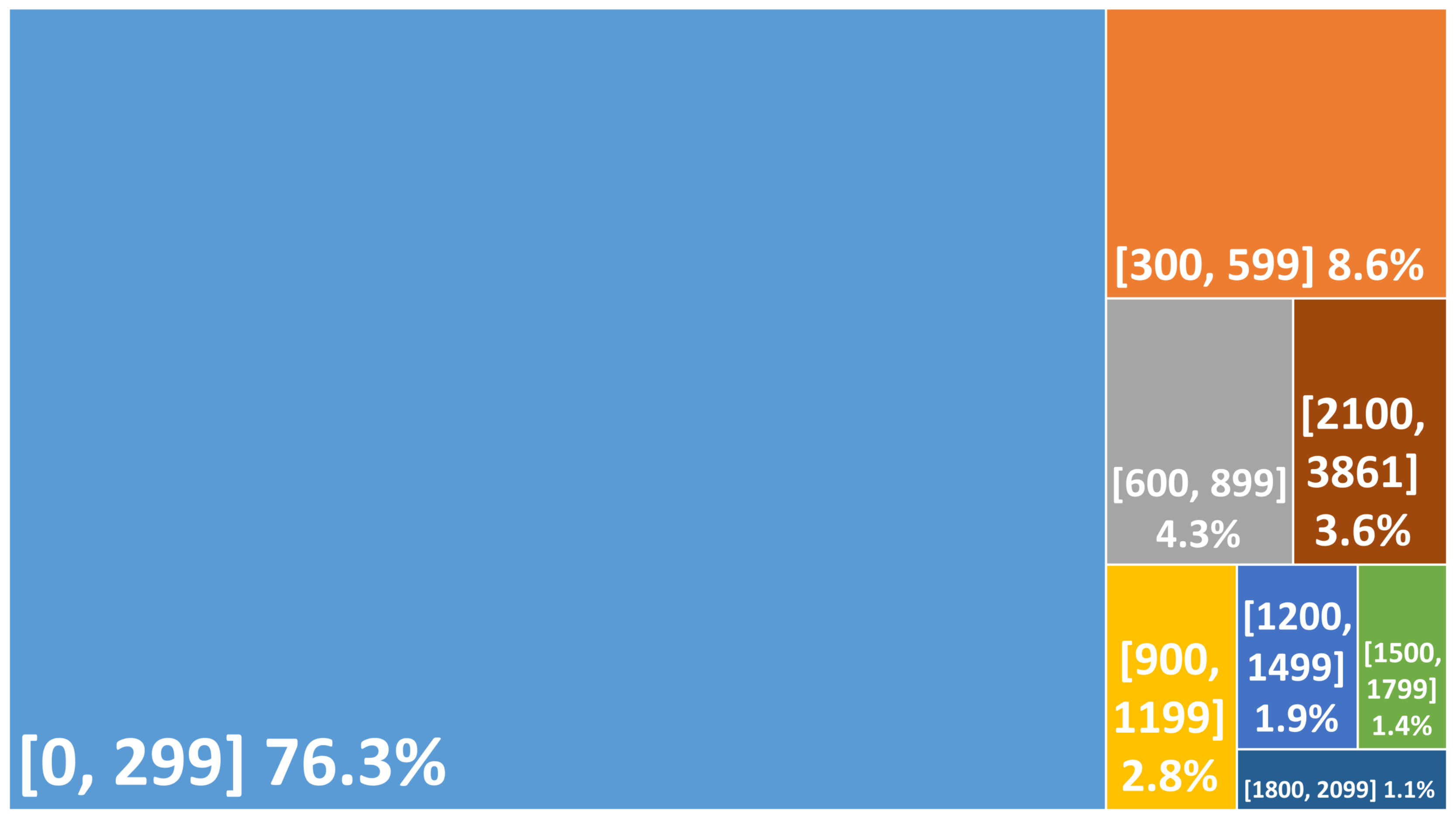}
\caption{\textbf{Visualization of training set imbalance.} The top 300 labels represent 76\% of the total multi-labels. Expanding the range to the top 900 labels takes up 89\% of the total.}
\label{fig:trainingimbalance}
\end{figure}

\begin{figure}
\centering
\begin{minipage}{.4\textwidth}
	\centering
	\includegraphics[height=.6\linewidth]{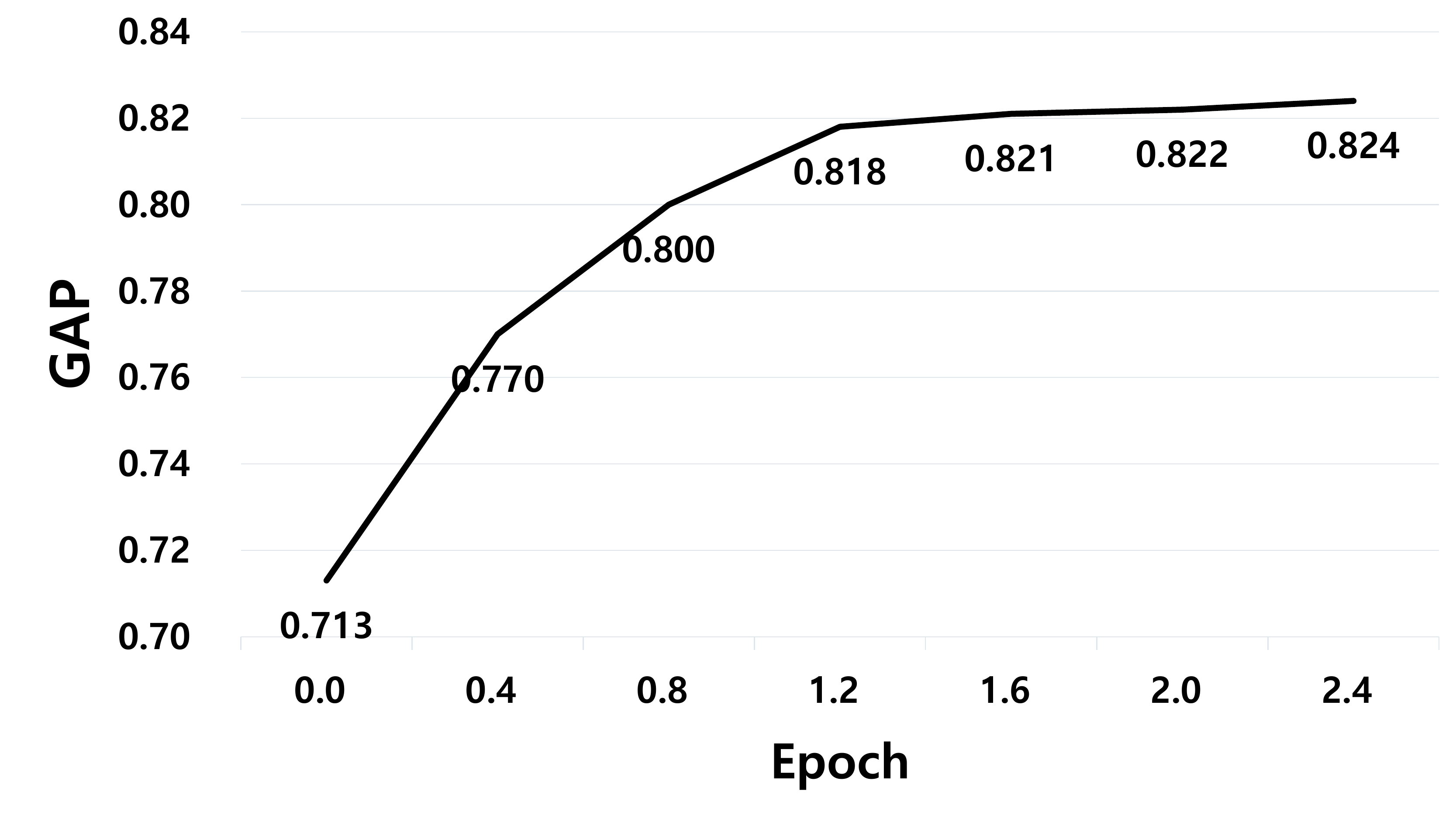}
	\caption{\textbf{GAP curve about validation set: Training with a small training set and retraining with the default training set.} 0.0 Epoch means when overfitted with the small training set.}
  	\label{fig:smalltofull}
\end{minipage}%
\begin{minipage}{.1\textwidth}
\quad
\end{minipage}%
\begin{minipage}{.4\textwidth}
	\centering
	\includegraphics[height=.6\linewidth]{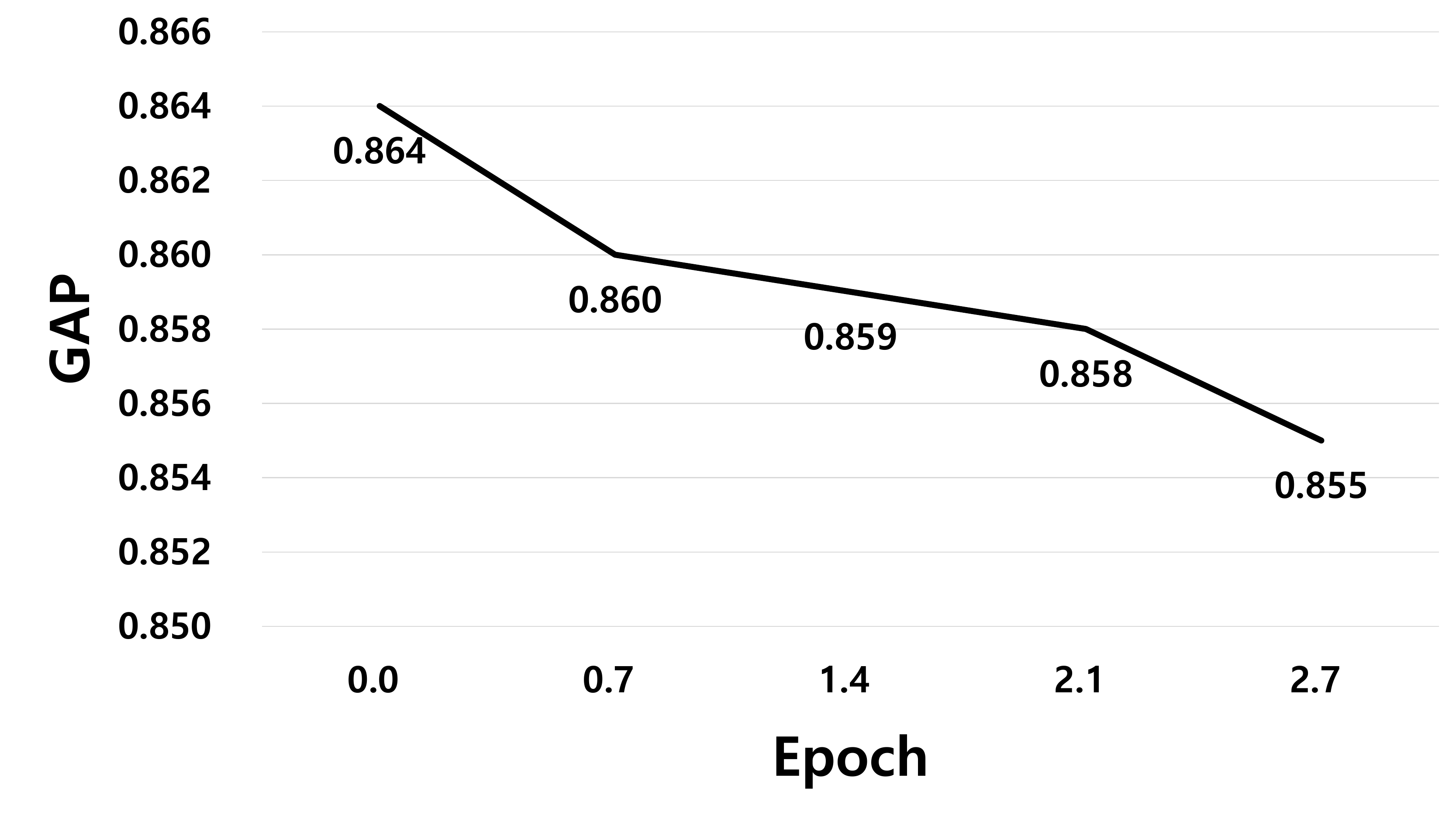}
	\caption{\textbf{GAP curve about validation set: Training with the default training set and retraining with the small training set.} 0.0 Epoch is when 2.5 epochs were trained with the default training set.}
  	\label{fig:fullandsmall}
\end{minipage}
\end{figure}

Fig.~\ref{fig:trainingimbalance} illustrates that the top 900 labels in the training set accounted for 89\% of the multi-label video data, or 10,445,267 of the 11,711,620 total labels for video data.
The remaining 2,962 labels represent only 10\% of 1,266,353 individuals.
To solve this data imbalance, a small training set was constructed with a label index > 977.
In this small training set, one epoch of training is performed at 7,300 steps with 2-GPU and each 80-batch (160-batch in total).
This small training set was used in two ways.
The first method was to train the small training set when the train GAP converged to 1.0 and retrain it as the existing default training set.
However, this performance was lower than that of the existing GAP of 0.86 (see Fig.~\ref{fig:smalltofull}).
In the second method, 2.5 epochs were trained with the default training set and retrained with a small training set (see Fig.~\ref{fig:fullandsmall}).
But, performance dropped when trained with a small training set.

\subsubsection{Analysis of Validation Results and Additional Experiment}
The correct answer was compared to the top 20 prediction results of the model.
The model validation GAP is 0.86.
The analysis showed that 117,410 in a total of 1,112,357 validation set did not include some or all of the correct answers in the top 20 predictions.
Of these, 21,828 were single-label, and 55,470 had more than 4 labels. 
Those with 1 label involved a unique feature of the video label, and those with 4 or more labels had overlapping label features and did not train well.
So, training data was selected with only 1 or 4-plus labels.
These data were added to the training data by tripling them from other data.
In all, about 8,500,000 large training set was created and trained.
Unfortunately, there was no significant performance improvement; with a difference of only about 0.0001 according to the GAP, no performance improvement was found through dataset preprocessing.
A clearer interpretation method is needed.

\subsection{Final Submission Model}
\quad
An optimal hyperparameter for the NetVLAD and NetFV models was found through the above methods. The results of the test set are shown in Table~\ref{table:result}. 

\setlength{\tabcolsep}{4pt}
\begin{table}
\begin{center}
\caption{\textbf{The performance (GAP) of each model.} Optimal cluster size, hidden size and GAP are shown.}
\label{table:result}
\begin{tabular}{llll}
\hline\noalign{\smallskip}
Model & Cluster size & Hidden size & GAP \\
\noalign{\smallskip}
\hline
\noalign{\smallskip}
NetVLAD & 192 & 1,200 & 0.86668 \\
NetFV & 120 & 1,024 & 0.86633 \\
\noalign{\smallskip}
\hline
\noalign{\smallskip}
Result & & & {\bf 0.86668} \\
\hline\noalign{\smallskip}
\end{tabular}
\end{center}
\end{table}
\setlength{\tabcolsep}{1.4pt}

\quad
In the end, a GAP score of 0.8668 was obtained at the 2nd YouTube-8M video understanding challenge.

\section{Conclusions}
\quad
This paper used video classification of the YouTube-8M dataset, applying the NetVLAD and NetFV models with reference to previous research data and using the Huber loss function.
Experimental verification is effective for improving performance by adjusting the training epoch, learning rate, and training set.
Unlike in conventional training for classification problem, the performance of 2.5 epochs is found to be optimal, as the training set is sufficiently large.
The learning rate was also adjusted for optimal training.
Even though no performance improvement was found, an attempt was made to train with the set that emphasized frequently-wrong patterns.

\subsubsection{Acknowledgement}
\quad
This work was partly supported by Institute for Information \& communications Technology Promotion(IITP) grant funded by the Korea government(MSIT) (2017-0-01772, Development of QA system for video story understanding to pass Video Turing Test), Information \& communications Technology Promotion(IITP) grant funded by the Korea government(MSIT) (2017-0-01781, Data Collection and Automatic Tuning System Development for the Video Understanding), and Institute for Information \& communications Technology Promotion(IITP) grant funded by the Korea government(MSIT) (No.2017-0-00271, Development of Archive Solution and Content Management Platform)


\bibliographystyle{splncs04}
\bibliography{egbib}

%Author, F.: Article title. Journal \textbf{2}(5), 99--110 (2016)
%Author, F., Author, S.: Title of a proceedings paper. In: Editor,
%F., Editor, S. (eds.) CONFERENCE 2016, LNCS, vol. 9999, pp. 1--13.
%Springer, Heidelberg (2016). \doi{10.10007/1234567890}
%Author, F., Author, S., Author, T.: Book title. 2nd edn. Publisher,
%Author, A.-B.: Contribution title. In: 9th International Proceedings
%on Proceedings, pp. 1--2. Publisher, Location (2010)
%LNCS Homepage, \url{http://www.springer.com/lncs}. Last accessed 4
%\end{thebibliography}

\end{document}